\documentclass[conference,a4paper]{IEEEtran}
\IEEEoverridecommandlockouts
\ifCLASSINFOpdf
\else
\fi
\usepackage{cite}
\usepackage{amsmath,amssymb,amsfonts}
\usepackage{booktabs} 
\usepackage{graphicx}
\usepackage{textcomp}
\usepackage{multirow} 
\usepackage{subcaption}
\def\BibTeX{{\rm B\kern-.05em{\sc i\kern-.025em b}\kern-.08em
    T\kern-.1667em\lower.7ex\hbox{E}\kern-.125emX}}

\usepackage[usenames,dvipsnames]{color}
\usepackage[square,numbers,sort]{natbib}
\usepackage[hyperfootnotes=false]{hyperref}
\definecolor{lightblue}{rgb}{0.1, 0.1, 1.0}
\hypersetup{
 colorlinks=True,
 citecolor=lightblue,
 linkcolor=lightblue,
 urlcolor=Blue
 }
\usepackage[utf8]{inputenc}
\usepackage{algorithm}
\usepackage{algpseudocode}
\usepackage{caption}

\begin{document}

\title{Color Enhancement for V-PCC Compressed Point Cloud via 2D Attribute Map Optimization\\}

\author{\author{
  \IEEEauthorblockN{Jingwei Bao\IEEEauthorrefmark{1}, Yu Liu\IEEEauthorrefmark{1}, Zeliang Li\IEEEauthorrefmark{2}, Shuyuan Zhu\IEEEauthorrefmark{1}, Siu-Kei Au Yeung\IEEEauthorrefmark{2}}
  \IEEEauthorblockA{\IEEEauthorrefmark{1}School of Information and Communication Engineering, University of Electronic Science and Technology of China}
  \IEEEauthorblockA{\IEEEauthorrefmark{2}School of Science and Technology, Hong Kong Metropolitan University}
  \IEEEauthorblockA{\{jwbao, yuliu16\}@std.uestc.edu.cn, s1218826@live.hkmu.edu.hk, eezsy@uestc.edu.cn, jauyeung@hkmu.edu.hk}
}

}

\maketitle

\begin{abstract}
Video-based point cloud compression (V-PCC) converts the dynamic point cloud data into video sequences using traditional video codecs for efficient encoding. However, this lossy compression scheme introduces artifacts that degrade the color attributes of the data. This paper introduces a framework designed to enhance the color quality in the V-PCC compressed point clouds. We propose the lightweight de-compression Unet (LDC-Unet), a 2D neural network, to optimize the projection maps generated during V-PCC encoding. The optimized 2D maps will then be back-projected to the 3D space to enhance the corresponding point cloud attributes. Additionally, we introduce a transfer learning strategy and develop a customized natural image dataset for the initial training. The model was then fine-tuned using the projection maps of the compressed point clouds. The whole strategy effectively addresses the scarcity of point cloud training data. Our experiments, conducted on the public 8i voxelized full bodies long sequences (8iVSLF) dataset, demonstrate the effectiveness of our proposed method in improving the color quality.
\end{abstract}
\begin{IEEEkeywords}
Point cloud compression, Image restoration, Transfer learning, Point cloud reconstruction
\end{IEEEkeywords}
\IEEEpeerreviewmaketitle
\section{Introduction}
A point cloud represents three-dimensional data as a collection of discrete points, each with coordinates and their corresponding attributes such as vertex normals and color. This data format is increasingly used in various fields such as virtual/augmented reality (VR/AR) creation, robotics, urban modelling, and autonomous vehicles, highlighting its growing importance \cite{tulvan2016use}. The large volume of point cloud data presents significant data management challenges. To address these, the moving picture experts group (MPEG) has developed two compression standards: geometry-based point cloud compression (G-PCC) for statistic point cloud and video-based point cloud compression (V-PCC) for dynamic point cloud \cite{graziosi2020overview}.

\begin{figure}[htbp]
\vspace{-0.5cm}
\centerline{\includegraphics[width=\linewidth]{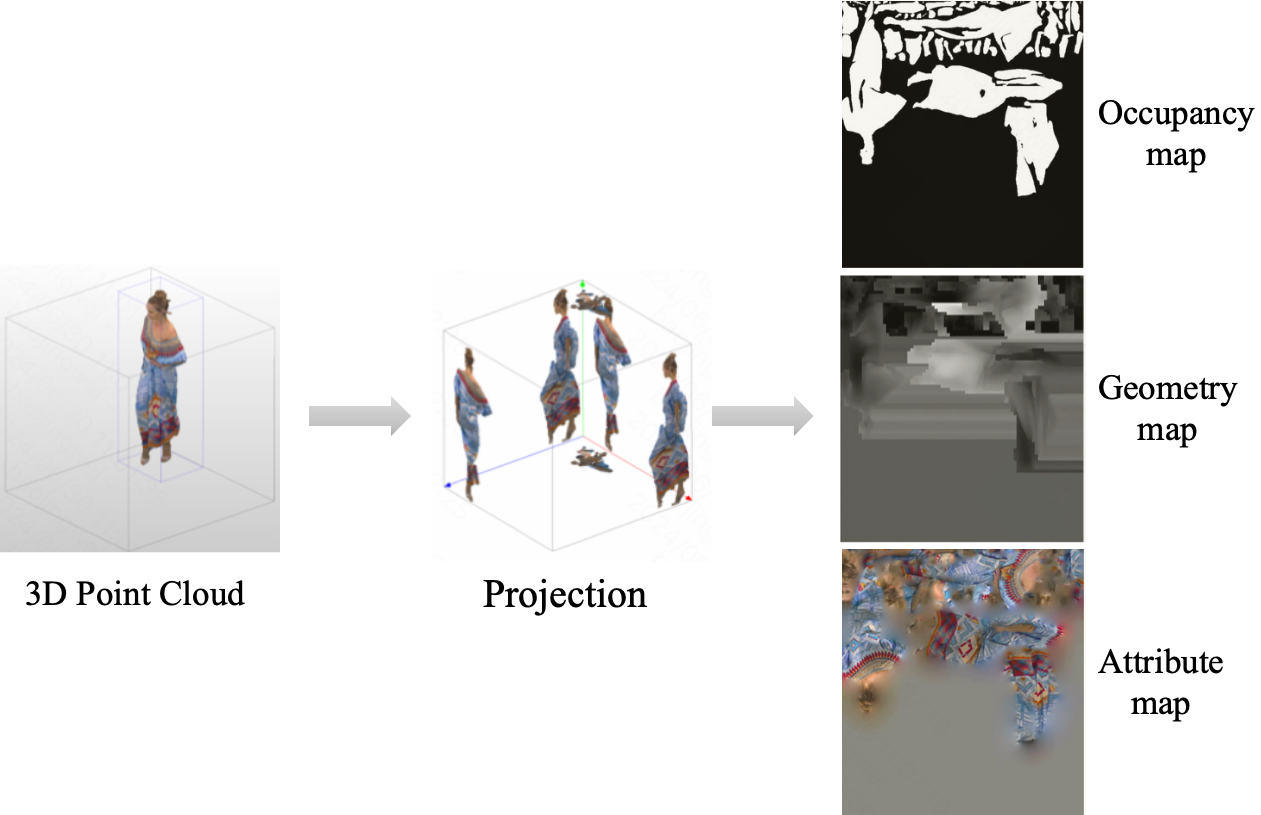}}
\caption{Projection in V-PCC.}
\label{fig1}
\vspace{-0.5cm}
\end{figure}

V-PCC utilizes video encoding technologies, including H.264 \cite{wiegand2003overview}, H.265 \cite{sullivan2012overview} and others, to compress the point cloud data by transforming it into 2D video streams. This process creates three types of frames: occupancy maps, which indicate valid 3D projection points; geometry maps, which provide the depth information; and attribute maps, which contain the color information of the points as shown in Fig. \ref{fig1}. Leveraging video compression technology, V-PCC achieves high compression rates and low latency, facilitating efficient point cloud data storage and transmission.

Preserving color fidelity is crucial for maintaining the visual integrity of compressed point clouds. Inevitably, lossy compression distorts color, particularly under low bit rate conditions. Current methodologies focus on mitigating color distortion and reducing artifacts. Li et al. introduced a novel method using the Minkowski Engine in a sparse, fully convolutional network to refine color information in V-PCC compressed point clouds affected by geometric distortions \cite{Li2023Sparse}. Similarly, Gao et al. developed the OCARNet model, employing a joint attention mechanism with MMSE loss to enhance color reconstruction in lossy-compressed point clouds characterized by both geometric and color degradation \cite{10416804}. Despite these advances, the previously proposed methods do not adequately control for the interference from geometric compression when addressing color distortions in lossy compressed point clouds. A more pressing challenge lies in the reliance on limited and hard-to-acquire three-dimensional point cloud data for training, which restricts the full potential of these models. Particularly, methods such as sparse convolutional neural networks, which are 3D deep learning techniques, are heavily dependent on substantial data volumes and pose training challenges due to the intrinsic complexity of extracting features in 3D spaces.

In this paper, we fully take into account the characteristics of V-PCC and propose a method based on the optimization of the projected attribute maps to enhance the color information of the compressed point clouds. The main contributions are summarized as follows:

\begin{itemize}
\item We proposed a framework to enhance point cloud attributes, using 2D methods to optimize the color information of V-PCC compressed point clouds, addressing the issuses of the lack of public datasets and the high complexity of model training.

\item We developed LDC-Unet, leveraging the U-Net architecture \cite{ronneberger2015u}, as a lightweight neural model to reduce parameters and  remove compression artifacts for attribute maps.

\item We constructed a task-associated dataset comprising a two-dimensional natural image collection of human portraits. This dataset, constructed based on attribute map processing during V-PCC encoding and data content similarity, is used to train the LDC-Unet.

\item We designed a transfer learning strategy. This involves training the model on the custom 2D portrait image dataset and fine-tuning it with point cloud projection maps to enhance generalization.
\end{itemize}

The rest of this article is organized as follows. We first introduce the proposed V-PCC compressed point cloud quality enhancement framework from three aspects: dataset construction, the architecture of the model, and transfer learning strategy design in Section \ref{sec:framework}. We then describe the configuration of the experimental and analyze the optimized results from both subjective and objective perspectives in Section \ref{sec:experiments}. Finally, we provide the conclusion of the paper in Section \ref{sec:conclusion}.
\section{PROPOSED METHOD}
\label{sec:framework}
In this section, we present our framework for enhancing attribute quality in V-PCC compressed point clouds. As depicted in Fig. \ref{fig3}, we first create a training dataset from a large collection of natural portrait images, which closely resemble the content of the point clouds. This dataset, comprising compressed and original images with corresponding masks, is used to train our LDC-Unet model. We then utilize a few V-PCC compressed point cloud projection maps for transfer learning. After training, we input the test point clouds’ occupancy maps and attribute maps, produced by V-PCC lossy compression, into the model for optimization. The optimized point clouds with improved color information are subsequently reconstructed in 3D by the decoder. The process is summarized as follows:
\begin{figure*}[t]
  \centering
  \resizebox{0.9\textwidth}{!}{\includegraphics{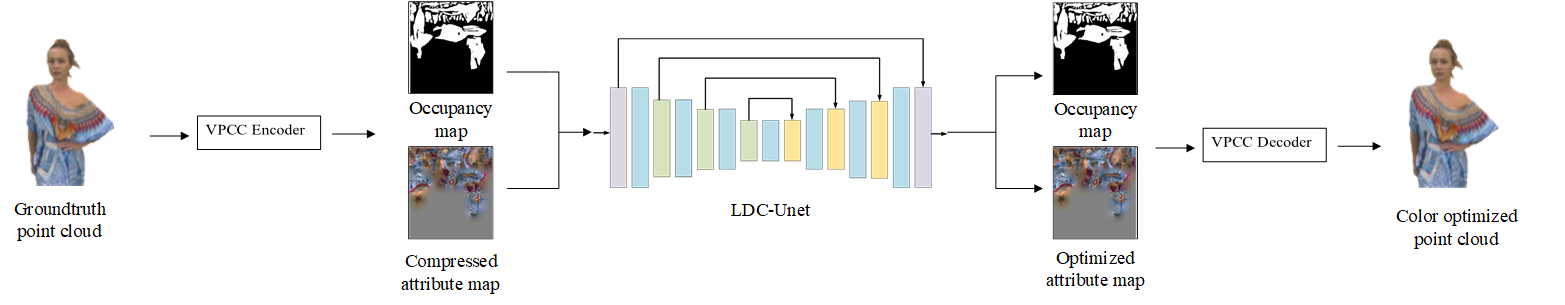}}
  \caption{V-PCC compressed point cloud color information enhancement framework} 
  \label{fig3} 
\end{figure*}
\begin{align}
y' &= \textit{M}(y) = \textit{M}(C(\textit{P}(x))) \label{eq1} \\
x' &= \textit{P}^{-1}(y') \label{eq2}
\end{align}
\( y \) represents the compressed attribute maps, and \( x \) denotes the original point cloud. The 3D-to-2D projection process used in encoding is denoted by \( P \), while \( C \) signifies the lossy compression. The model's two-dimensional optimization processing is represented by \( M \), resulting in \( y' \), the quality-enhanced attribute map. The 2D-to-3D back projection is denoted by \( P^{-1} \), leading to \( x' \), the attribute-enhanced reconstructed point cloud.

\subsection{Dataset construction}
\begin{figure}[h]
\vspace{-0.2cm}
\centerline{\includegraphics[width=\linewidth]{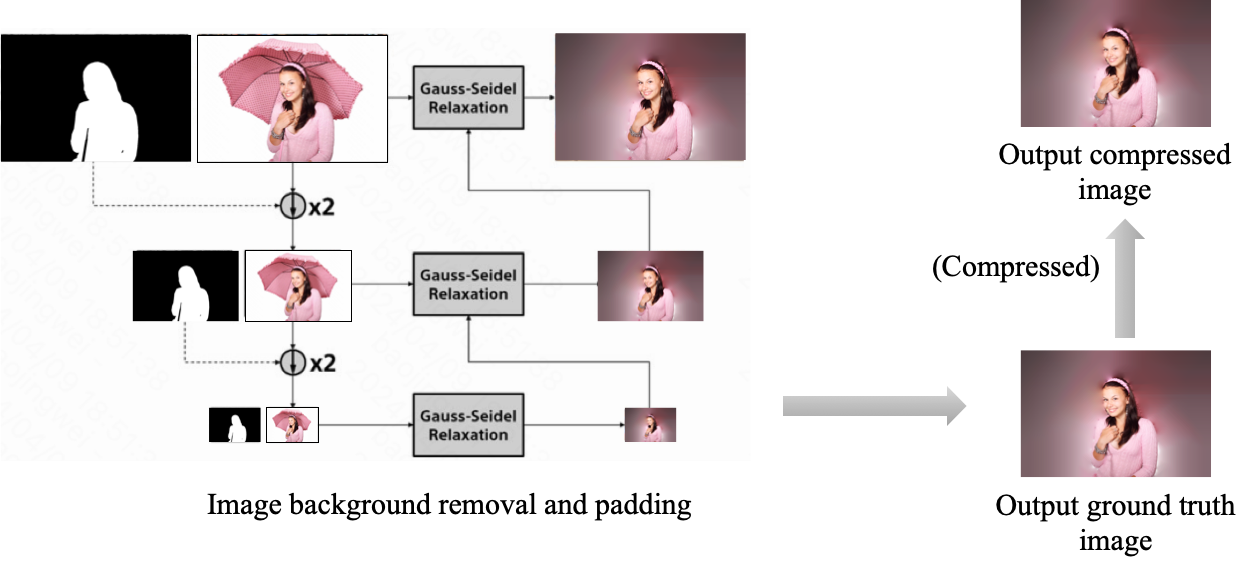}}
\caption{2D images dataset preparation}
\label{fig4}
\vspace{-0.2cm}
\end{figure}

Our point cloud optimization framework is based on optimization through 2D projection maps, enabling the use of abundant and easily accessible natural image data for the training rather than being limited to using only scarce point cloud data. We selected the publicly accessible supervisely persons dataset \cite{SuperviselyPersons} for our training set. This dataset includes 2246 human scene images along with the corresponding masks. Utilizing these, a dataset was prepared for training on the optimization of point cloud attribute maps as illustrated in Fig. \ref{fig4}. Initially, masks were used to remove non-human backgrounds from the images. An algorithm for background filling, akin to that employed in V-PCC and leveraging mip-map interpolation, was further refined through sparse linear optimization to generate the ground truth images \cite{graziosi2020overview}. Finally, these images were compressed using HEVC \cite{sullivan2012overview}, with the quantization parameter (QP) aligned with those used in the V-PCC, to produce noisy inputs for our model. The masks and compressed images can be compared to the occupancy maps and attribute maps generated by V-PCC encoding, demonstrating significant similarities.
\begin{figure}[ht]
\vspace{-0.2cm}
\centerline{\includegraphics[width=\linewidth]{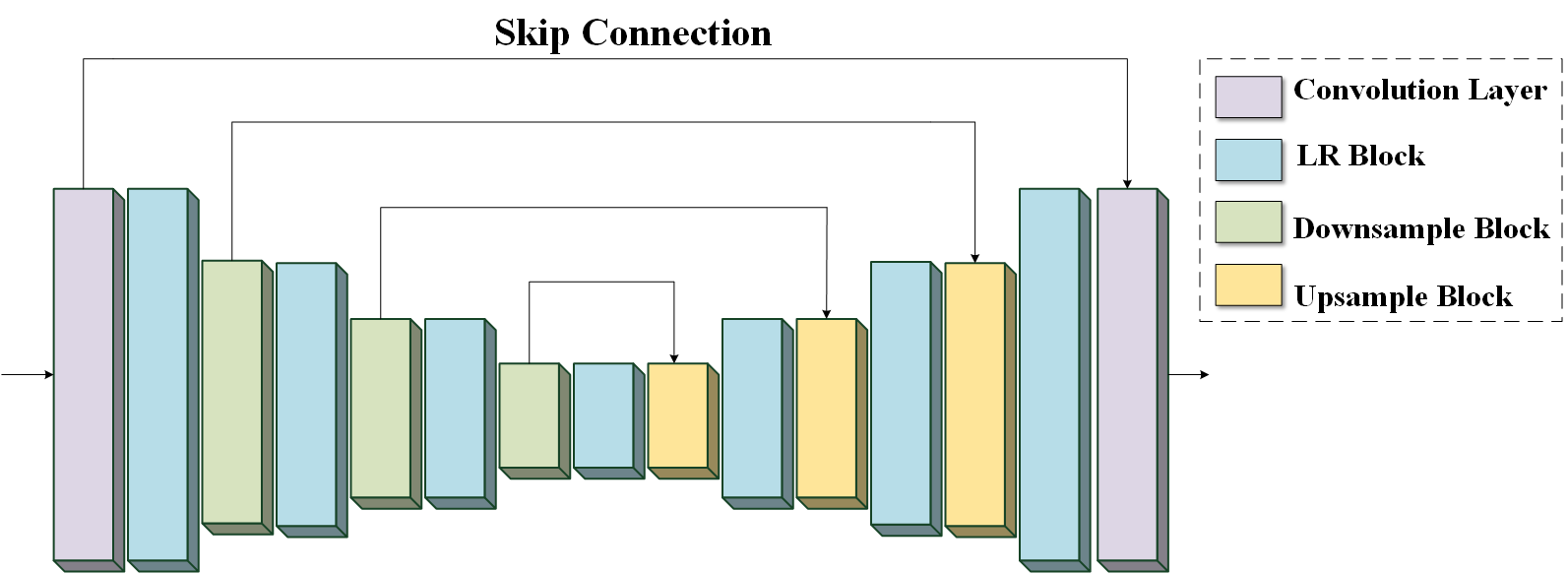}}
\caption{The architecture of LDC-Unet}
\label{fig5}
\vspace{-0.2cm}
\end{figure}
\begin{figure}[htbp]
\vspace{-0.5cm}
\centerline{\includegraphics[width=1\linewidth]{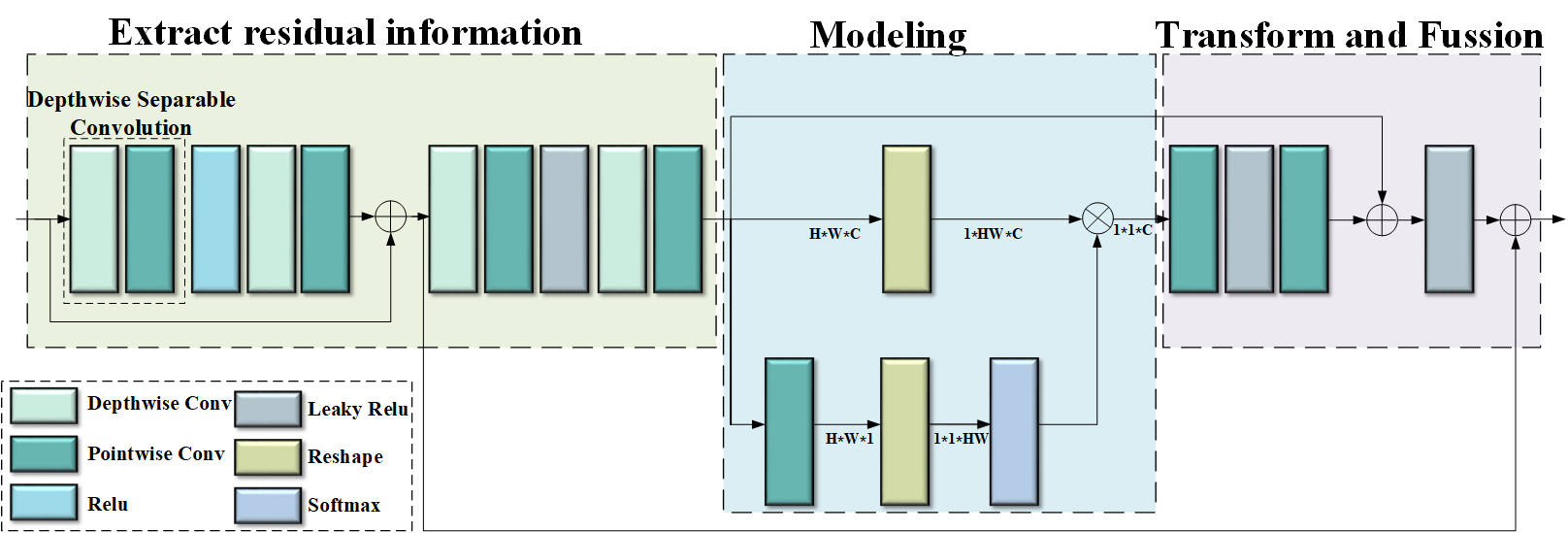}}
\caption{The architecture of LR block}
\label{fig6}
\vspace{-0.5cm}
\end{figure}
Another dataset is derived from a small batch of point cloud data, Waterloo Point Cloud sub-dataset (WPCSD) \cite{liu2021pqa}. We scaled and augmented this point cloud dataset, resulting in a total of 75 point cloud files. These files were then processed through the V-PCC encoder for projection and compression, from which occupancy maps and attribute maps were extracted. These maps will be utilized in subsequent transfer learning processes.
\subsection{Architecture of the model}
Our network architecture is shown in Fig. \ref{fig5}, which is inspired by DRUNet \cite{zhang2021plug}. DRUNet combines U-Net with residual connections, integrating residual blocks at each downsampling stage to enhance gradient flow and feature extraction. Skip connections address the vanishing gradient problem by enabling direct gradient backpropagation pathways, while residual blocks improve feature learning effectiveness and mitigate overfitting. Residual blocks also refine multi-scale feature integration within the U-Net architecture, enhancing the network’s capability to capture and fuse information across diverse scales. However, the incorporation of residual blocks in DRUNet, particularly as network depth increases, leads to a significant increase in parameter count. Therefore, we explore more efficient implementations that maintain optimal utilization of residual information while ensuring overall model lightweightness. We designed a lightweight residual block (LR block), leading to the development of a new, more efficient architecture, the LDC-Unet.

As shown in Fig. \ref{fig6}, within the LR block, the number of residual blocks per scale change was halved from four, and the depthwise separable convolution (DSC) \cite{howard2017mobilenets} was utilized in lieu of standard convolutions. DSC reduces parameters and enhances efficiency by decomposing convolution into depthwise (DConv) and pointwise (PConv) operations:
\begin{align}
\text{DConv:} & \quad \mathbf{Y}_{\text{depth}} = \mathbf{X} * \mathbf{K}\\
\text{PConv:} & \quad \mathbf{Y} = \mathbf{Y}_{\text{depth}} \cdot \mathbf{P}
\end{align}
where $\mathbf{X} \in \mathbb{R}^{H \times W \times C_{\text{in}}}$ is the input feature map, $\mathbf{K} \in \mathbb{R}^{k \times k \times C_{\text{in}}}$ and $\mathbf{P} \in \mathbb{R}^{C_{\text{out}} \times C_{\text{in}}}$ are depthwise and pointwise convolution kernels, respectively. By sharing parameters across channels and reducing dimensionality, DSC achieves efficiency ideal for lightweight models and resource-constrained environments. Subsequently, the residual information is subjected to context modelling to derive a global feature description, followed by feature transformation and fusion. This aspect of the design is inspired by the residual contextual block (RCB) \cite{zamir2022learning}. The LR block not only reduces the parameter counts but also suppresses less valuable features within the residual information, allowing the passage of features with more substantial information, thereby enhancing the model's efficiency.

\subsection{Transfer learning training strategy}
To overcome the generalization limitations commonly encountered by models using 3D point cloud data, such as sparse fully convolutional networks \cite{Li2023Sparse}, we designed a novel transfer learning strategy. The theoretical advantage of transfer learning lies in its ability to leverage knowledge and experience acquired from previous learning tasks to enhance the learning process and performance of a target task, particularly when data is scarce or distributions differ. Our approach leverages the abundant and trainable natural image data to construct an initial training dataset, which facilitates effective knowledge transfer to the point cloud projection maps subsequently.

In the first phase of training, we input processed natural images and their respective masks into the model, using L1 loss to guide the compressed images towards their ground truth. The masks function as binary maps similar to occupancy maps, providing structural information, while the processed and compressed natural images mimic quantization errors akin to those in V-PCC.

In the second phase, we employ a small batch of point cloud data for V-PCC encoding and projection to derive occupancy and attribute maps. These maps are subsequently utilized to refine the initial model. Our 2D model, employing a transfer learning strategy, offers a streamlined training approach and enhanced performance compared to 3D neural network models.
\section{EXPERIMENTS}
\label{sec:experiments}
In this section, we detail the experimental procedures and analyze both the objective and subjective outcomes. This comprehensive evaluation showcases the enhancements achieved in V-PCC compressed point cloud attributes through the proposed methods.
\subsection{Dataset and Training Parameters}
In our experiments, we selected the first 32 frames of the "\textit{soldier}","\textit{longdress}","\textit{loot}", and "\textit{redandblack}" sequences from the dynamic point cloud dataset provided by 8i \cite{Krivokuca2018} for testing. We compressed these using the V-PCC's TMC2 version 18 \cite{VPCC2022} encoder with standard rate configurations of r1 to r3 (attribute QP of 42, 37, 32) and performed lossless compression on the geometry data. We focused on enhancing the quality of the Y channel in the point clouds' YCbCr color space and measured the PSNR in both 2D and 3D. In the initial training phase, we set both the batch size and the number of epochs to 30. Upon entering the second phase, we reduced the number of epochs to 10. The initial learning rate was set at $1 \times 10^{-4}$ and was halved during the transfer learning phase. All experiments were conducted on a system equipped with an NVIDIA GeForce Titan GPU. We utilized the trained model to optimize compressed attribute maps generated during the V-PCC compression process on the test set point clouds, using the optimized results for 3D reconstruction to produce restored point clouds.


\subsection{Experimental Results}
In our experiments, we used the official MPEG point cloud quality measurement software \cite{Tian2017} to measure the model's PSNR results on the Y channel of both attribute maps and reconstructed point clouds across different sequences and compression parameters in the test set, and demonstrated subjective effects. Additionally, we confirmed that our network model maintains performance while reducing parameter count compared to DRUNet.

Table \ref{tab1} demonstrates the optimization effect of the Y channel in the attribute map of compressed point clouds in the model. Phase 1 represents the initial stage of training the model using a dataset consisting of human images, while Phase 2 showcases the results after transfer learning. The table clearly illustrates that LDC-Unet significantly enhances color information quality in 2D, as transfer learning has been proven effective. The final column displays the ultimate optimization effect after Phase 2.

Table \ref{tab2} presents the final optimization effect of the model on the 3D reconstructed point clouds. It can be observed from the table that, due to the model's ability to learn rich degradation information more easily at higher QP values, the optimization effect becomes more pronounced as the compression level increases.

Table \ref{tab3} presents a comparison of the training effectiveness between LDC-Unet and DRUNet. Both models were trained using the same dataset (portrait dataset at QP42), and their optimization effects on attribute maps from the redandblack and soldier sequence compressed at QP42 using V-PCC were tested. The results demonstrate that while reducing the parameter count, LDC-Unet maintains a similar model performance to that of DRUNet.

\begin{figure}[ht]

    \centering
    \begin{subfigure}[b]{0.3\linewidth}
        \centering
        \includegraphics[width=\textwidth]{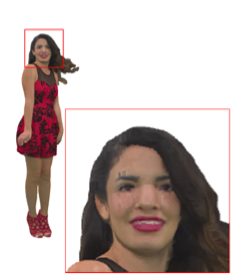}
        {\scriptsize \centerline{Source point cloud}}
        \label{fig7:sub1}
    \end{subfigure}
    \begin{subfigure}[b]{0.3\linewidth}
        \centering
        \includegraphics[width=\textwidth]{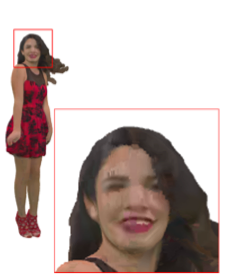}
        {\scriptsize \centerline{Compressed point cloud}}
        \label{fig7:sub2}
    \end{subfigure}
    \begin{subfigure}[b]{0.3\linewidth}
        \centering
        \includegraphics[width=\textwidth]{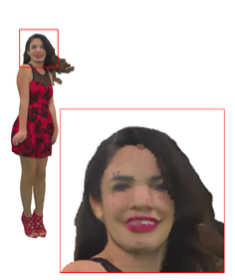}
        {\scriptsize \centerline{Enhanced point cloud}}
        \label{fig7:sub3}
    \end{subfigure}
    \caption{Subjective evaluation}
    \label{fig7}

\end{figure}

Fig. \ref{fig7} shows the subjective visual effects of the first frame point clouds of the "\textit{redandblack}" sequences at QP42, after model optimization. It can be observed that the color blurring in the facial regions is significantly alleviated following color optimization, demonstrating the subjective effectiveness of our method.
\section{Conclusion}
\label{sec:conclusion}
This paper introduces a novel framework leveraging the LDC-Unet model and transfer learning with a task-specific dataset to enhance attribute quality in V-PCC compressed point clouds, particularly focusing on improving color fidelity. However, the study is limited by its exclusive optimization of color attributes and the assumption of lossless compression for geometry to prevent interference. Future work aims to address these limitations by optimizing geometry maps under lossy V-PCC compression conditions using a specialized two-dimensional dataset. This approach promises to enhance geometric quality in reconstructed point clouds, potentially broadening the applications of point cloud compression technology.
\newpage
\begin{table}[htbp]
  \centering
  \caption{Color Enhancement Result for 2D Attribute Maps}
  \label{tab1}
  \setlength{\tabcolsep}{4pt}
  \renewcommand{\arraystretch}{1.2}
  \begin{tabular}{@{}c|c|ccc@{}c}
  \toprule
    \multicolumn{2}{c}{} & \multicolumn{3}{c}{PSNR(dB) on Y Channel} & \\
  \midrule
  Test Sequences & QP & Noisy Input & Phase 1 & Phase 2   & \multicolumn{1}{c}{Improvements}\\ \hline
                 & 42 & 32.7182     & 32.9732 & \textbf{33.1706}    & 0.4524 \\
  \textit{soldier}        & 37 & 35.3269     & 35.6397 & \textbf{35.7596}    & 0.4327 \\
                 & 32 & 38.1508     & 38.4324 & \textbf{38.5572}    & 0.4064 \\ \hline
                 & 42 & 32.4163     & 32.7147 & \textbf{32.8013}    & 0.3850 \\
  \textit{longdress}      & 37 & 35.1046     & 35.3702 & \textbf{35.4136}    & 0.3090 \\
                 & 32 & 37.9874     & 38.2205 & \textbf{38.2562}    & 0.2688 \\ \hline
                 & 42 & 36.2375     & 36.5294 & \textbf{36.7022}   & 0.4647\\
   \textit{loot} & 37 & 38.6453     & 38.8921 & \textbf{39.0111}   & 0.3658 \\
                 & 32 & 41.3614     & 41.6027 & \textbf{41.7402}   & 0.3788 \\ \hline 
                 & 42 & 37.2328     & 37.7052 & \textbf{37.9431}   & 0.7103\\
    \textit{redandblack}  & 37 & 39.3645     & 39.7480 & \textbf{39.9095}   & 0.5450 \\
                 & 32 & 41.7409     & 42.0651 & \textbf{42.2195}   & 0.4786 \\
  \bottomrule
  \end{tabular}
\end{table}
\begin{table}[H]

  \centering
  \caption{Color Enhancement Result for 3D Reconstructed Point Clouds}
  \label{tab2}
  \setlength{\tabcolsep}{5pt} 
  \renewcommand{\arraystretch}{1.2} 
  \begin{tabular}{@{}c|c|ccc@{}}
  \toprule
    \multicolumn{2}{c}{}                & \multicolumn{3}{c}{PSNR(dB) on Y Channel} \\
  \midrule
  Test Sequences & QP & Input & Final Output& Improvements\\ \hline
         & 42 & 29.5187 & 29.7254 & 0.2067   \\
  \textit{soldier}               & 37 & 31.9788 & 32.1467 & 0.1679   \\
                 & 32 & 34.6734 & 34.8008 & 0.1274   \\ \hline
       & 42 & 27.5597 & 27.7556 & 0.1959   \\
   \textit{longdress}              & 37 & 30.1076 & 30.2410 & 0.1334   \\
                 & 32 & 32.8601 & 32.9431 & 0.0830\\ \hline
             & 42 & 32.1752 & 32.2596 & 0.0844  \\
   \textit{loot}              & 37 & 34.4427 & 34.4687 & 0.0260\\
                 & 32 & 37.0737 & 37.0964 & 0.0227  \\ \hline 
      & 42 & 32.4408 & 32.6397 & 0.1989  \\
    \textit{redandblack}             & 37 & 34.3294 & 34.4597 & 0.1303  \\
                 & 32 & 36.5341 & 36.6016 & 0.0675  \\
  \bottomrule
  \end{tabular}
\end{table}
\begin{table}[H]

\centering
\caption{Comparison of 2D models}
\label{tab3}
\begin{tabular}{c|c|c|c}
\hline
                           & Parameters & \textit{Soldier} & \textit{redandblack} \\ \hline
DRUNet                      & 32638656   & 32.9560   & 37.6425   \\ \hline
LDC-Unet                     & 4035136   & 32.9732   & 37.7052     \\ \hline
\end{tabular}

\end{table}


\newpage 
\bibliographystyle{IEEEtran}
\bibliography{refer}

\begin{thebibliography}{10}
\providecommand{\url}[1]{#1}
\csname url@samestyle\endcsname
\providecommand{\newblock}{\relax}
\providecommand{\bibinfo}[2]{#2}
\providecommand{\BIBentrySTDinterwordspacing}{\spaceskip=0pt\relax}
\providecommand{\BIBentryALTinterwordstretchfactor}{4}
\providecommand{\BIBentryALTinterwordspacing}{\spaceskip=\fontdimen2\font plus
\BIBentryALTinterwordstretchfactor\fontdimen3\font minus \fontdimen4\font\relax}
\providecommand{\BIBforeignlanguage}[2]{{%
\expandafter\ifx\csname l@#1\endcsname\relax
\typeout{** WARNING: IEEEtran.bst: No hyphenation pattern has been}%
\typeout{** loaded for the language `#1'. Using the pattern for}%
\typeout{** the default language instead.}%
\else
\language=\csname l@#1\endcsname
\fi
#2}}
\providecommand{\BIBdecl}{\relax}
\BIBdecl

\bibitem{tulvan2016use}
C.~Tulvan, R.~Mekuria, Z.~Li, and S.~Laserre, ``Use cases for point cloud compression (pcc),'' ISO/IEC JTC1/SC29/WG11 (MPEG) output document N16331, 2016.

\bibitem{ronneberger2015u}
O.~Ronneberger, P.~Fischer, and T.~Brox, ``U-net: Convolutional networks for biomedical image segmentation,'' in \emph{Medical Image Computing and Computer-Assisted Intervention--MICCAI 2015: 18th International Conference, Munich, Germany, October 5-9, 2015, Proceedings, Part III 18}.\hskip 1em plus 0.5em minus 0.4em\relax Springer, 2015, pp. 234--241.

\bibitem{graziosi2020overview}
D.~Graziosi, O.~Nakagami, S.~Kuma, A.~Zaghetto, T.~Suzuki, and A.~Tabatabai, ``An overview of ongoing point cloud compression standardization activities: Video-based ({V-PCC}) and geometry-based ({G-PCC}),'' \emph{APSIPA Transactions on Signal and Information Processing}, vol.~9, p. e13, 2020.

\bibitem{wiegand2003overview}
T.~Wiegand, G.~J. Sullivan, G.~Bj{\o}ntegaard, and A.~Luthra, ``Overview of the {H.264/AVC} video coding standard,'' \emph{IEEE Transactions on Circuits and Systems for Video Technology}, vol.~13, no.~7, pp. 560--576, 2003.

\bibitem{sullivan2012overview}
G.~J. Sullivan, J.-R. Ohm, W.-J. Han, and T.~Wiegand, ``Overview of the high efficiency video coding {(HEVC)} standard,'' \emph{IEEE Transactions on circuits and systems for video technology}, vol.~22, no.~12, pp. 1649--1668, 2012.

\bibitem{VPCC2022}
``{V-PCC Test Model v18},'' {ISO/IEC JTC1/SC29/WG07 MPEG/N00311, Online}, Apr. 2022.

\bibitem{sheng2022attribute}
X.~Sheng, L.~Li, D.~Liu, and Z.~Xiong, ``Attribute artifacts removal for geometry-based point cloud compression,'' \emph{IEEE Transactions on Image Processing}, vol.~31, pp. 3399--3413, 2022.

\bibitem{xing2023gqe}
J.~Xing, H.~Yuan, R.~Hamzaoui, H.~Liu, and J.~Hou, ``{GQE-Net}: a graph-based quality enhancement network for point cloud color attribute,'' \emph{IEEE Transactions on Image Processing}, vol.~32, pp. 6303--6317, 2023.

\bibitem{Li2023Sparse}
Z.~Li, J.~Bao, Y.~Liu, S.-K.~A. Yeung, S.~Zhu, K.~Hung, and M.~A. Khan, ``Sparse fully convolutional network for video-based point cloud compression color enhancement,'' in \emph{Proc. 2023 6th Artificial Intelligence and Cloud Computing Conference (AICCC 2023)}.\hskip 1em plus 0.5em minus 0.4em\relax Kyoto, Japan: ACM, Dec. 2023, pp. 1--8.

\bibitem{liu2021pqa}
Q.~Liu, H.~Yuan, H.~Su, H.~Liu, Y.~Wang, H.~Yang, and J.~Hou, ``Pqa-net: Deep no reference point cloud quality assessment via multi-view projection,'' \emph{IEEE transactions on circuits and systems for video technology}, vol.~31, no.~12, pp. 4645--4660, 2021.

\bibitem{zhang2021plug}
K.~Zhang, Y.~Li, W.~Zuo, L.~Zhang, L.~Van~Gool, and R.~Timofte, ``Plug-and-play image restoration with deep denoiser prior,'' \emph{IEEE Transactions on Pattern Analysis and Machine Intelligence}, vol.~44, no.~10, pp. 6360--6376, 2021.

\bibitem{zamir2022learning}
S.~W. Zamir, A.~Arora, S.~Khan, M.~Hayat, F.~S. Khan, M.-H. Yang, and L.~Shao, ``Learning enriched features for fast image restoration and enhancement,'' \emph{IEEE transactions on pattern analysis and machine intelligence}, vol.~45, no.~2, pp. 1934--1948, 2022.

\bibitem{Krivokuca2018}
M.~Krivokuća, P.~A. Chou, and P.~Savill, ``8i voxelized surface light field (8ivslf) dataset,'' ISO/IEC JTC1/SC29 WG11 (MPEG) input document m42914, Ljubljana, Jul. 2018.

\bibitem{SuperviselyPersons}
\BIBentryALTinterwordspacing
Supervisely, ``Supervisely persons,'' 2020. [Online]. Available: \url{https://ecosystem.supervisely.com/projects/persons}
\BIBentrySTDinterwordspacing

\bibitem{Tian2017}
D.~Tian, H.~Ochimizu, C.~Feng, R.~Cohen, and A.~Vetro, ``Updates and integration of evaluation metric software for pcc,'' MPEG input document M40522, 2017.

\bibitem{howard2017mobilenets}
A.~G. Howard, M.~Zhu, B.~Chen, D.~Kalenichenko, W.~Wang, T.~Weyand, M.~Andreetto, and H.~Adam, ``Mobilenets: Efficient convolutional neural networks for mobile vision applications,'' \emph{arXiv preprint arXiv:1704.04861}, 2017.

\bibitem{10416804}
L.~Gao, Z.~Li, L.~Hou, Y.~Xu, and J.~Sun, ``Occupancy-assisted attribute artifact reduction for video-based point cloud compression,'' \emph{IEEE Transactions on Broadcasting}, pp. 1--14, 2024.

\end{thebibliography}

\end{document}